# A machine learning method for the large-scale evaluation of urban visual environment

Lun Liu[a]    Hui Wang[b*]    Chunyang Wu[c]
a Department of Land Economy, University of Cambridge
b School of Architecture, Tsinghua University
c Machine Intelligence Laboratory, Department of Engineering, University of Cambridge

**Abstract**

Given the size of modern cities in the urbanising age, it is beyond the perceptual capacity of most people to develop a good knowledge about the beauty and ugliness of the city at every street corner. Correspondingly, for planners, it is also difficult to accurately answer questions like 'where are the worst-looking places in the city that regeneration should give first consideration', or 'in the fast urbanising cities, how is the city appearance changing', etc. To address this issue, we here present a computer vision method for the large-scale and automatic evaluation of the urban visual environment, by leveraging state-of-the-art machine learning techniques and the wide-coverage street view images. From the various factors that are at work, we choose two key features, the visual quality of street façade and the continuity of street wall, as the starting point of this line of analysis. In order to test the validity of this method, we further compare the machine ratings with ratings collected on site from 752 passers-by on fifty-six locations. We show that the machine learning model can produce a good estimation of people's real visual experience, and it holds much potential for various tasks in terms of urban design evaluation, culture identification, etc.

## 1 Introduction

As a city grows large, it becomes hardly possible for its dwellers, as well as planners, to gather a complete knowledge about how it looks, at every street corner and in every narrow alley. Theoretically, the human perception of urban environment is inherently incomplete, discontinuous and distorted, as depicted by research on cognitive mapping (Downs and Stea 1973) and the city's image (Lynch 1960), especially given the overwhelming size of modern cities. It makes questions hard to answer such as 'where are the worst-looking places in the city that regeneration should give first consideration', or 'in fast urbanising cities, how is the city appearance changing', etc.

Actually, there have long been attempts in measuring city's appearance in a consistent manner and in a larger scale (Harvey 2014). The dominant method is by sending human auditors to the field to observe and record (Brownson, Hoehner et al. 2009), but still, this method is quite limited in sample size since its manual nature makes it inherently expensive and derive few economy of scale (Harvey 2014). Recently, the availability of online street view images, which has an unprecedentedly wide coverage on the built environment, provides a new methodological opportunity into this topic (Hara, Le et al. 2013; Kelly, Wilson et al. 2013; Hwang and Sampson 2014). When combined with computer vision techniques, there is a possibility for the large-scale automatic evaluation of various high-level judgements on the urban built environment (Doersch, Singh et al. 2012; Salesses, Schechtner et al. 2013; Naik, Philipoom et al. 2014; Ordonez and Berg 2014; Quercia, O'Hare et al. 2014; Lee, Maisonneuve et al. 2015).

Our goal in this paper is thus to explore this possibility in terms of the urban visual experience. We refer to architectural and urban design theories and choose two visual features, the visual quality of architecture facade and the visual continuity of street wall, as a starting point in this study. These two features are influential to the urban visual experience and reasonably explainable. We then use Beijing, a fast-growing city with quite diverse visual appearances, as the case of study.

However, the use of street view images and computer vision is challenged by several issues in producing a right estimation of people's real experience. First, we used expert rating to train the model. Despite of the theoretical soundness of the rating criteria, there may be a gap between experts' opinion and the public's preference. Second, the rating is based on piecemeal, static, 2-D images instead of the on-site, dynamic, 3-D experience. In particular, the image is incapable of demonstrating the lighting condition in the real setting, which creates ever-changing shade and shadow on a façade that gives various senses of visual depth, solidity and architectural beauty (Carmona 2010). Moreover, there is also difference between the angle, height and viewing distance of the street view camera and human vision, especially the fisheye lens effect of street view images, which may result in distorted judgements.

Regarding to this, there have been a few studies comparing the results of observational field audits and street view image-based audits and showing that there is generally an agreement (Hara, Le et al. 2013; Kelly, Wilson et al. 2013). However, what those studies dealt with are quite objective and straightforward factors such as the building height, obstacles on the sidewalk, etc., while the visual features we are looking at are integrated judgements. In order to test the validity of our proposed method, we further conducted a field survey on 752 passers-by at fifty-six locations in Beijing and compare the public's ratings with the computer ratings. The result demonstrates that this method can produce a good evaluation of people's real experience.

The main contributions of our paper are:
- developing machine learning models for the large-scale evaluation on the urban visual environment;
- validating the proposed models against the public's opinions collected from the field survey data;
- producing the Beijing visual environment evaluation map, from which various planning implications can be made.

The rest of the paper is organised as follows: Section 2 reviews the recent progress in applying machine learning on extracting high-level perceptual and cultural information from city images; Section 3 explains the theoretical base of the visual features modelled in this study; Section 4 introduces the data and methodology; Section 5 presents the performance of the machine learning model and the validation results, and the Beijing visual environment evaluation map produced from the model results; Section 6 concludes and discussed the potential directions of research.

## 2 Related works

Previously, most computer vision algorithms related to places have focused on technical tasks such as scene classification, or parsing scene images into constituent objects and background elements (Ordonez and Berg 2014). Building upon that, there have emerged a few interesting research into the perceptual and cultural aspects of urban images in recent years.

In the seminal work of '*What makes Paris look like Paris*', Doersch et al. dealt with the identification of local architectural identity by proposing a discriminative clustering approach that automatically discovers geographically representative elements from vast Google Street View images. They identified several visually interpretable and perceptually geo-informative architectural elements that distinguish Paris from other European and North American cities, including the floor-to-ceiling window with cast iron railings, the decorative balcony supports, the emblematic street signs, etc. (Doersch, Singh et al. 2012). Related to that, there is also a research line into the automatic classification of architectural styles by capturing the morphological characteristics, which can be further applied to the identification of architectural style mix and style transformation over time (Shalunts, Haxhimusa et al. 2011; Goel, Juneja et al. 2012; Shalunts, Haxhimusa et al. 2012; Xu, Tao et al. 2014; Lee, Maisonneuve et al. 2015).

The most relevant works to this paper are those that aim at understanding people's perception of urban scenes, which are usually analysed by crowd-sourcing ratings on urban images. Quercia et al. identified the aesthetic informative elements that positively (e.g. the amount of greenery) or negatively (e.g. broad streets, fortress-like buildings, etc.) affect people's perception of beauty, quietness and happiness (Quercia, O'Hare et al. 2014). Ordonez and Berg modelled the perception for wealth, uniqueness and safety judged from street view images and validated the results against ground-truth income and crime statistics (Ordonez and Berg 2014). The perception of safety is also modelled by Naik et al. and Porzi et al. (Naik, Philipoom et al. 2014; Porzi, Rota Bulò et al. 2015) and proved to be in consistency with actual socio-economic indicators (Naik, Kominers et al. 2015).

Our task in this paper is somewhat different by focusing on the visual instead of perceptual understanding of urban scenes. Our selected visual features are more directly linked to architectural and urban design, and thus can be easily translated into guidelines in planning practice. We here select two visual features as a starting point to this line of research. Many other features that have long been addressed in design codes may also be fed into machine learning in future research, such as architectural style, human scale, compatibility, consistency and contrast, etc.

# 3 Selection of visual features

## 3.1 Visual quality of architecture façade

The architecture façade is a highly influential component of the urban space that concentrates visual attention and 'radiates' onto the urban space (Von Meiss 2013). Actually, the urban space is, to a large extent, shaped by the building façades, which arrest the eyes and also the space that would otherwise slip by (Buchanan 1988). The visual quality of façade is a combined effect of various factors. It include but are not limited to:

- **Composition** which creates visual rhythms and holds the attention (Buchanan 1988). It is formed by the repetition of constituent parts (e.g. windows, doors, bays), the ratio of solid to void, the articulation of vertical and horizontal elements, etc. (Carmona 2010).
- **Material** which gives texture and pattern to the surface and applies a certain visual friction to slow the eye and space (Buchanan 1988).
- **Detail** which holds the eye and provides interest. A space can feel harsh and inhuman if its surfaces lack fine details and visual interest, while finely detailed, a space can be delicate, airy and inviting (Carmona 2010). However, overloaded details can also have a counter effect, since too much complexity is tested to be negatively correlated with people's preference (Devlin and Nasar 1989).
- **Color** which evokes feelings and emotions. According to Wassily Kandinsky, each color is linked to a certain feeling, such as red to alive, restless, blue to deep, inner, supernatural, peaceful, etc. (Kandinsky 2012).
- **Maintenance** which keeps the façade in a decent condition. Related issues include whether dirt and stains are regularly removed, whether deteriorated or damaged components are repaired, whether replacement materials or details match the original, etc.

## 3.2 Visual continuity of street wall

The street wall refers to the interface formed by building façades along a street. A continuous street wall is formed when buildings are lined in a row without significant interruptions caused by vacant lots or setbacks. A continuous street wall offers a sense of enclosure (Ewing and Clemente 2013), majesty and controlled uniformity (Milroy 2010), and draws pedestrians and activities. As early as the 15th century, relevant rules had appeared in a street design codes in Numberg, Germany, which required buildings to be lined up to create an "undeviating building line" (Kostof 1992). Nowadays, it is addressed in numerous planning codes and guidelines, e.g. the APA Planning and Urban Design Standards requires infill projects to "maintain ground floor façade to define a consistent street edge", and if there is any parking lots between buildings, the street wall should be continued "by means of an attractive fence, masonry wall, or hedges" (APA 2006).

# 4 Data and methodology

## 4.1 Case of study

We used Beijing as the case of study, which has undergone dramatic transformation from the imperial capital to the administrative center, and now also, to a hotspot of global investment. Such experience has turned the cityscape into a complex mosaic of traditional and super modern, and small and giant structures. Besides, its rapid expansion in the recent decade has resulted in considerable amount of poorly constructed buildings at the urban fringe, where the cityscape is much different from that of the city center. All in all, the current image of Beijing is no longer the so-called 'Old Beijing' composed of grey brick walled courtyards fronted by venerable red wooden doors in the maze of lanes. The highly diversified visual environment makes Beijing a vivid example for our analysis. We focused on the area within the 5$^{th}$ ring road, which covers most of the built-up areas.

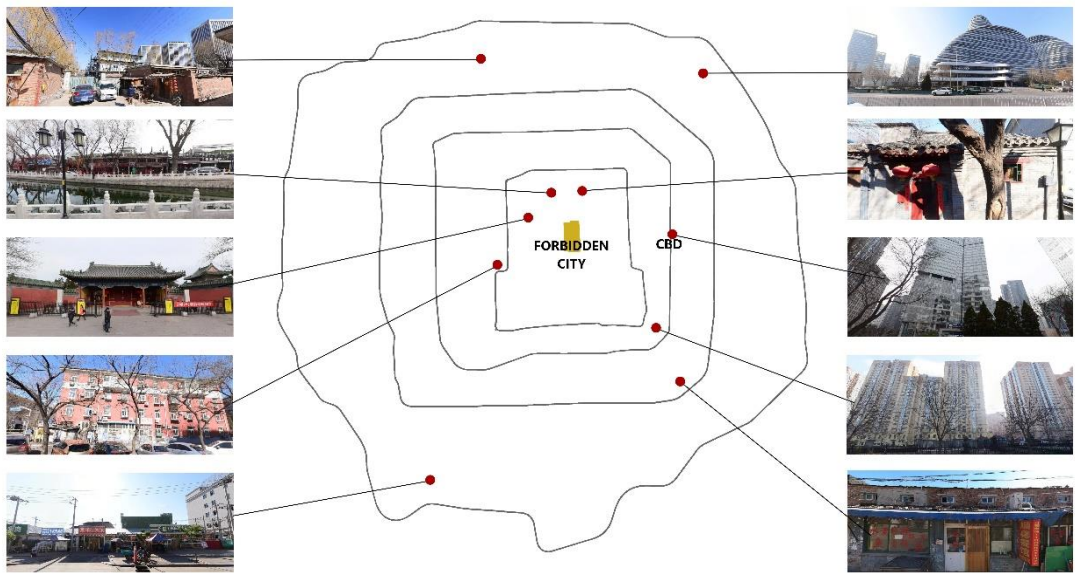

Figure 1 The research area and examples of the cityscape

We used street view images crawled from Baidu Map, the Chinese equivalent of Google Map. The images were requested at an interval of two hundred meters along all the streets in the city, returning 360,796 images (800*500 pixels). Different from most existing studies that focused on the entire streetscape and used images taken with the camera facing the street, we put more emphasis on the architectural façade and set the camera facing the buildings so that the architecture takes a larger proportion of the image (see Figure 2). However, about thirty percent of the images returned were still streetscape images, which are taken around street corners or entrances. Therefore, a machine learning model was developed to discern streetscape images from building images so that the unqualified can be screened out.

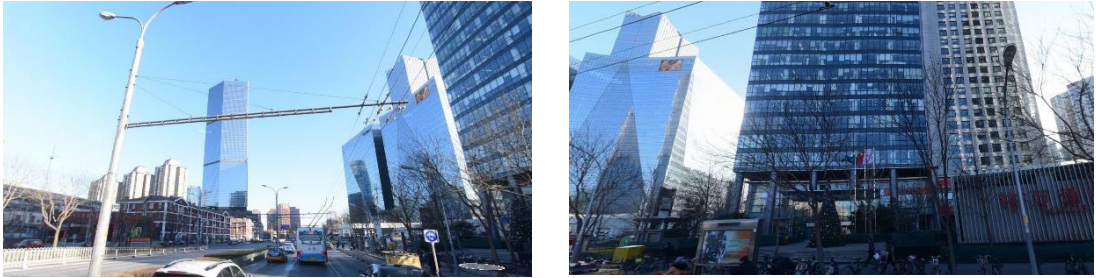
Figure 2 Camera facing the street (left) and facing the buildings (right)

## 4.2 Expert rating

We took a two-step approach in developing the machine learning models. In the first step, we randomly sampled 3,500 images from the database and manually labeled them as "building images" (2575) and "street images" (925) as shown in Figure 2. They were then used to train a "qualification" model to decide whether the content of an image is appropriate to be included in the analysis. In the next step, the qualified "building images" were manually rated on the two visual features by architectural students who have studied the subject for more than five years. The two ratings were then fed to develop the visual quality rating model and the visual continuity rating model. The rating criteria are explained below.

According to the review in Section 3.1, the visual quality of architecture façade is contributed by fine textured materials, good quality of details, appropriate coloring, rhythmic composition of solid and void, and vertical and horizontal elements, as well as regular maintenance. In the case of Beijing, we rate the visual quality into four classes from one point to four points. Four points are given to façades that meet almost all the standards above, which usually appear on newly built commercial houses, high-end office buildings, well maintained traditional architecture, etc. Three points are given to those that are less well-composed, designed with fewer details, use cheaper materials, and have not-so-pretty such as hanging wires and iron window fences. Nevertheless, this group of building façades generally present a neat and clean look. Many of them appear on the matchbox apartment blocks, which were mass-produced in the 1970s to 1990s and appear to be monotonous and repetitive. Those rated two points are built with hardly any rhythm or detail, and cheap materials. Besides, they are usually subject to inadequate maintenance, resulting in messy hanging wires, stained walls, exfoliated surfaces, rusty iron fences, etc. One point is given to those in a quite dilapidated condition, featured by very stained walls, ramshackle roofs, temporary building material such as metal roof sheet, etc., which usually happen at the urban fringe. The expert rating returns 485 four points images (18.8%), 1079 three points images (41.9%), 809 two points images (31.4%), and 202 one point images (7.8%).

Regarding to the continuity of street wall, we put an emphasis on the lower floors, since the horizontal field of vision gives weight to lower height that is reachable by our eyes (Gehl 2013). That's to say, if the façade is continuous at the lower floors but divided into separated blocks at higher floors, it is still considered to be continuous. If a solid wall is placed between the building and the street, usually in a gated community, then there is considered to be no architectural continuity since the building is blocked at the eye height. In the rating process, the images are classified into continuous and discontinuous types. The continuous refers to those that show building façades progressing through the image without any interruption, blockage or significant setback, at least at the eye height; otherwise an image is classified as discontinuous. The expert rating identifies 1069 "continuous" images (41.5%) and 1506 "discontinuous" images (58.5%).

Table 1 Distribution of expert rating

| Qualifiation | Qualified | Unqualified | | |
|---|---|---|---|---|
| | 73.6% | 26.4% | | |
| Visual quality | 4 Points | 3 Points | 2 Points | 1 Points |
| | 18.8% | 41.9% | 31.4% | 7.8% |
| Visual continuity | Continuous | Discontinuous | | |
| | 41.5% | 58.5% | | |

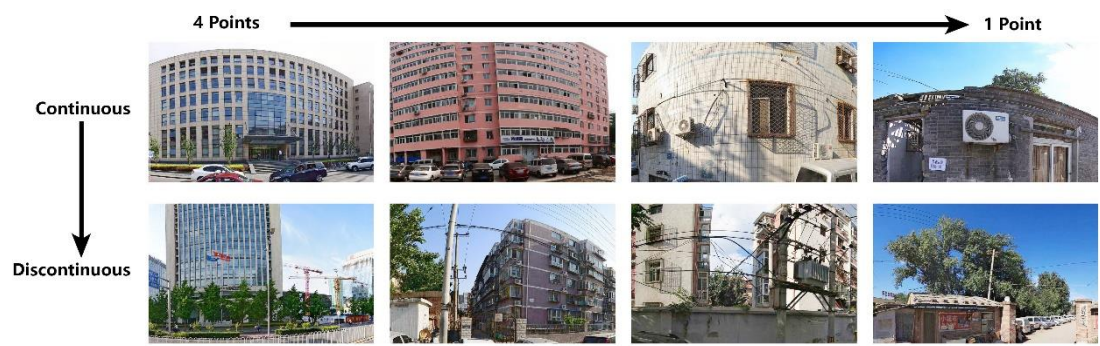

Figure 3 Rating examples

## 4.3 Machine learning

In the field of computer vision, there are quite a few approaches for image representation. For our work, we evaluate three features: the conventional SIFT histogram, and two state-of-the-art deep convolutional networks AlexNet and GoogLeNet, which outperform all other features in the 2012 and 2014 ImageNet Large Scale Visual Recognition Competition. Compare with conventional image techniques, which are dominated by low-level features like edges and corners, the deep convolutional networks are able to capture both local and high-level image characteristics. We use the output of the last hidden layer of the two pre-trained neural networks and train a SVM classifier for each of the scene attributes.

The labelled data set was divided into three subsets, the training set, the development set and the test set. For each task, the development set and the test set were equally and randomly sampled in each labelled class, and the rest of the images were used as the training set. For example, for the visual quality task, forty images were randomly sampled in each of the four scoring groups for the development set and sixty images each for the test set. We then adjusted the model parameters on the development set until the performance was optimized. In terms of the evaluation of model performance, we used the F1-score for the classification models (the qualification model and the visual continuity model) and the mean squared error (MSE) for the visual quality rating model, which are calculated as below:

$$Recall = \frac{TP}{P}$$

$$Precision = \frac{TP}{TP + FP}$$

$$F1 = \frac{2TP}{2TP + FN + FP} = \frac{2 \times Precision \times Recall}{Precision + Recall}$$

$$MSE = \frac{1}{n}\sum(y_i - t_i)^2$$

Where *P* (positive), *TP* (true positive), *FP* (false positive), *FN* (false negative) denote the number of the images that are qualified/continuous, both labelled and predicted to be qualified/continuous, labelled unqualified/discontinuous but predicted to be qualified/continuous, labelled true but predicted to be false, and $y_i$ and $t_i$ denote the machine rating and expert rating for each image.

The models with the best performance for the three tasks were chosen to be applied to the entire image database of the research area. We then calculated the average visual quality and visual continuity scores of each street segment and produced the Beijing visual environment evaluation map.

## 4.4 Validation survey

As mentioned in the introduction, in order to test the validity of the proposed method, we conducted a field survey to collect the public's opinion on their visual experience (on a scale of 1 to 5) and compared the results with the machine rating. The survey was conducted on fifty-six street segments, which included both low-rated and high-rated, traditional-looking and modern-looking streets. On each street segment, we randomly sampled ten to fifteen interviewees and at the same time, kept the demographic profile in consistency with the whole city (Table 1). The total sample size was 752 interviewees. For validation, the spearman's r was calculated for the machine ratings and the average score rated by the interviewees on the fifty-six street segments, which could show how well the machine rating represents the on-site rating.

Table 2 Descriptive statistics of the interviewees

| Variables | Frequency | %Share |
|---|---|---|
| *Gender* | | |
|     Male | 377 | 50.13% |
|     Female | 375 | 49.87% |
| *Age* | | |
|     <18 | 62 | 8.24% |
|     18-40 | 272 | 36.17% |
|     41-60 | 254 | 33.78% |
|     60+ | 164 | 21.81% |
| *Residence* | | |
|     Beijing resident | 249 | 33.11% |
|     visitor | 503 | 66.89% |
| *Education* | | |
|     Elementary school and under | 39 | 5.19% |
|     Junior school | 177 | 23.54% |
|     High school and equivalent | 267 | 35.50% |
|     Bachelor's degree and equivalent | 252 | 33.51% |
|     Master's degree and above | 17 | 2.26% |
| *Total* | 752 | 100% |

# 5 Results

## 5.1 Machine learning performance

Table 3 shows the performance of the SIFT, AlexNet and GoogLeNet features on the test set of the qualification task. The deep convolutional networks, the AlexNet and the GoogLeNet, performed better than the traditional SIFT features. The GoogLeNet achieved a slightly higher F1 score than the AlexNet, which indicated a more balanced performance between recall and precision. Table 4 shows the performance of the visual quality task. Similar to that on the qualification task, deep features outperformed the SIFT features. The GoogLeNet showed a better capability of generalisation with a lower MSE on the development set. Since the GoogLeNet has been proved to be better on the first two tasks, we directly applied it on the visual continuity task and the results are shown in Table 5.

Table 3 Performance of the qualification model

| | Accuracy (%) | Precision (%) | Recall (%) | F1 (%) |
|---|---|---|---|---|
| SIFTHist + SVM | 79.22 | 45.06 | 71.28 | 55.22 |
| AlexNet + SVM | 89.25 | **48.23** | 85.94 | 61.78 |
| GoogLeNet + SVM | **90.04** | 48.13 | **86.31** | **61.79** |

Table 4 Performance of the visual quality model

| MSE | Training set | Development set | Test set |
|---|---|---|---|
| SiftHist + SVM | 0.358 | 0.841 | 0.835 |
| AlexNet + SVM | **0.218** | 0.635 | **0.619** |
| GoogLeNet + SVM | 0.278 | **0.614** | 0.641 |

Table 5 Performance of the visual continuity model

| GoogLeNet + SVM | Accuracy (%) | Precision (%) | Recall (%) | F1 (%) |
|---|---|---|---|---|
| Training set | 82.38 | 41.57 | 76.18 | 53.79 |
| Development set | 75.00 | 46.00 | 69.00 | 55.20 |
| Test set | 75.00 | 48.00 | 72.00 | 57.60 |

In terms of the validation, the distribution of machine scores and survey scores for the sample street segments are shown below. It is worth noting that a large proportion of streets in Beijing are not enclosed by a continuous street wall, so the scores of visual continuity do not follow a normal distribution but skew towards zero. Besides, although the quality of Beijing's cityscape is widely criticized in the academia, the scores rated by the interviewees were mostly higher than the midpoint of the rating scale, which to some extent, indicates that the public has been used to the current situation and become less criticising. The correlation analysis showed that the machine scores and survey scores were moderately-to-highly correlated for both visual features (spearman's r = 0.66 for visual quality, 0.71 for visual continuity), which indicated that the machine learning algorithm can provide a good approximation to the public's visual experience in the real urban environment.

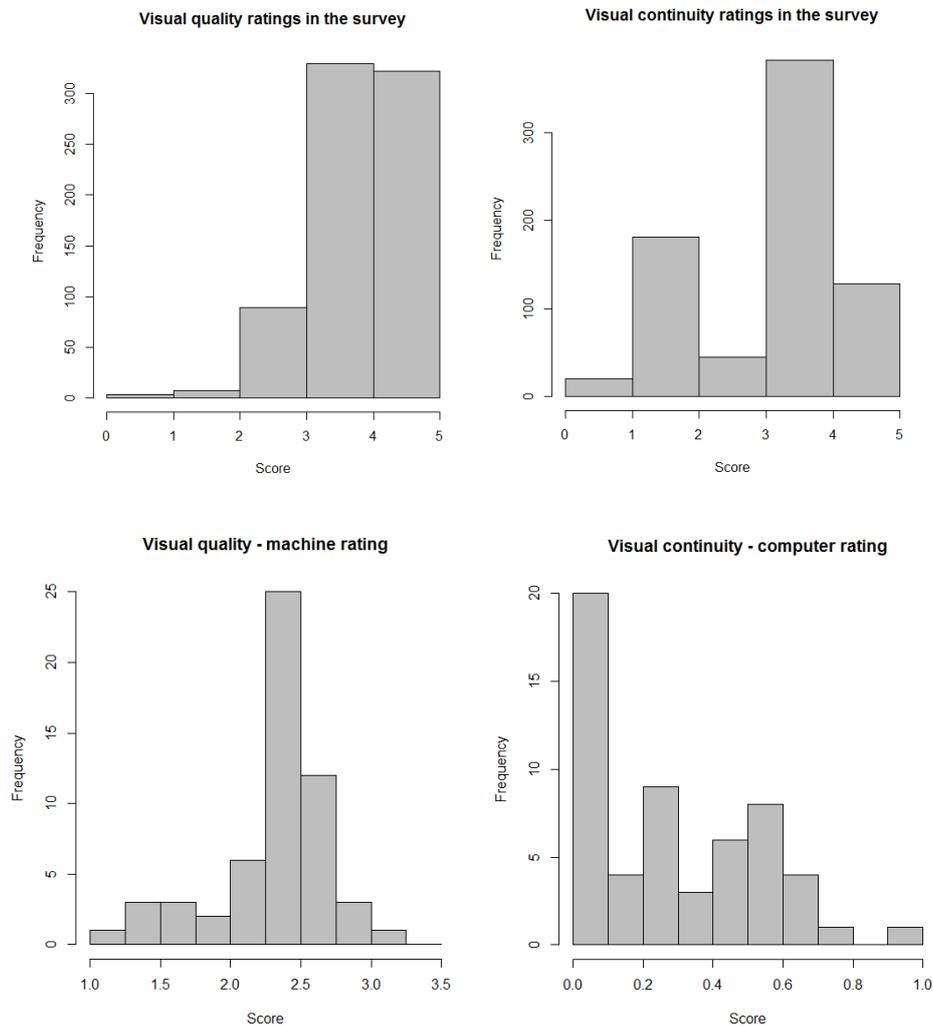

Figure 4 Distribution of scores

## 5.2 Beijing visual environment evaluation map

By calculating the average score of each street segment, we produced the scoring maps for the two visual features (Figure 5 and Figure 6). They provides lots of information about the visual environment, which can help with planning and urban design decision making. However, it needs to be noted that, although the model is technically valid, the scores should not be taken as absolutely accurate but an estimation with errors. For instance, the red colored street segments may not always be of higher visual quality than the orange ones, but are in most cases of higher quality than the blue ones. The purpose of this section is not to discuss every detail of the two maps, but present a few examples of the planning implications that can be drawn from this work.

We here take the whole city, the major avenues, and the blocks as the three levels of analysis, from which different types of patterns can be identified. For instance, in terms of the visual quality, there is apparent pattern at the city scale that the north part of the city (Zone A) generally scores higher than the south part (Zone B), especially at the urban fringes between the 4[th] and the 5[th] ring roads. When checked with the street view image, it shows that while most areas between the North 4[th]

Ring Road and the North 5th Ring Road maintain a modern urban look, many areas in the south resemble more of a dilapidated village than a city. Therefore, more urban planning and design effort is needed in the south city to fill in this gap. Regarding to the major avenues, for example, it draws attention that the north-south central axis (Zone C), which is considered to be the heritage of the ancient city and is given great importance, does not seem to present an outstanding architectural visual quality. Instead, the west-east axis (Zone D) appears to be more visually appealing. It therefore indicates the need of more measures to be taken in the making of this central axis. At the block-level, small concentrations of high scores and low scores can be identified. For instance, most key development areas score above the average and form a hotspot of warm colors on the map, which proves the success of place making in these areas, such as the CBD (Zone E), the second CBD (Zone F) and Zhongguancun IT Science Park (Zone G). However, some of them appears to be isolated with the surrounding areas since there is a sudden drop of score in the surroundings. For example, Zone H is a business park with decently designed office buildings, but on the other side of the adjacent railway lies shabby village houses (Zone I). Such imbalance in the development of the built environment also needs to be alleviated in planning practice.

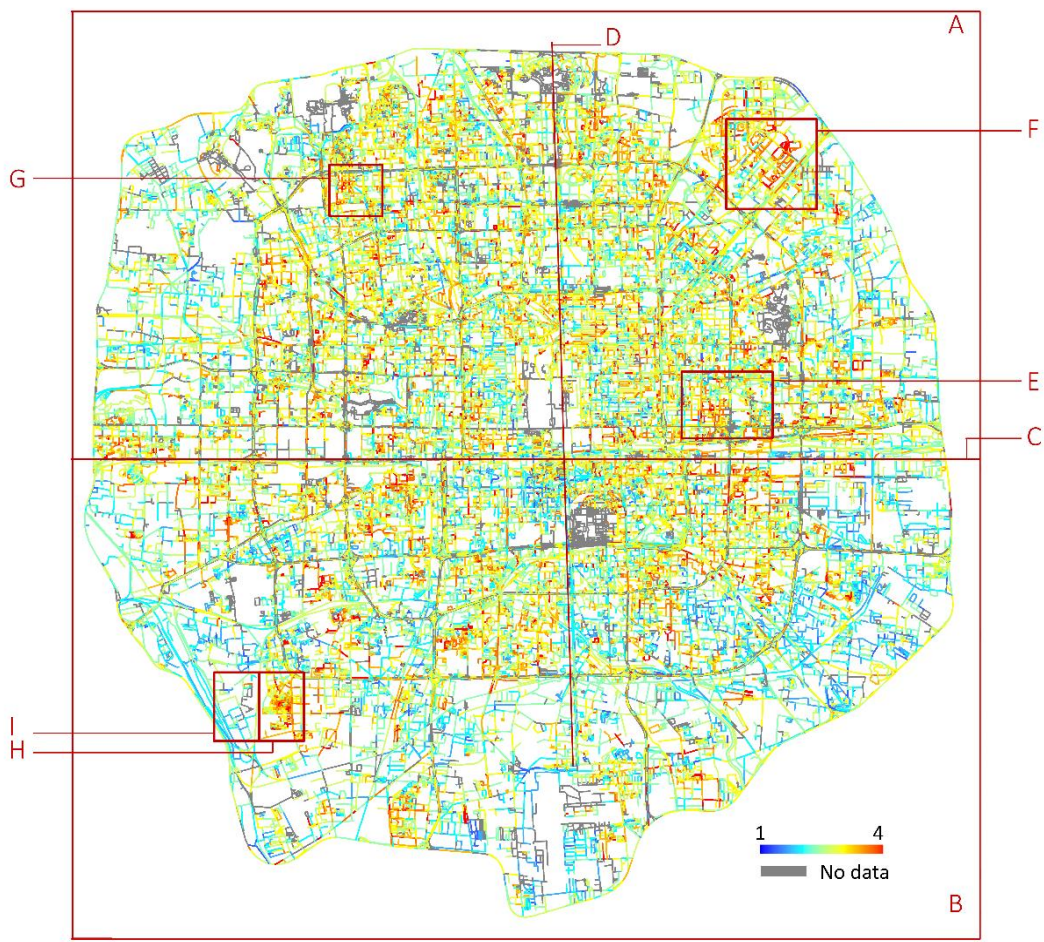

Figure 5 Visual quality scoring of architecture façade in Beijing

In term of the visual continuity rating, at the city scale, it is apparent that a large proportion of the street walls in Beijing are not continuous. The historical areas within the 2nd ring road (Zone J), where the streets take the form of traditional Hutongs, scores much higher than elsewhere in the city, which reveals the one of the major differences between the visual environment in the historical areas and modern developments. It reminds that, if the city is going to keep its urban identity, not only the architectural styles, but also this kind of structural feature needs to be preserved and inherited. Besides, the street walls along the ring roads are generally more continuous, since they are considered to be the gateways of the city and the streetscape is given more emphasis. Regarding to the key development areas, they turn out to be much less outstanding in Figure 6 than in Figure 5, which indicates that the high quality individual buildings fail to provide a feeling of continuity as a whole. In summary, the visual continuity rating demonstrates the need to incentivize infill development and more aggressively regulate shallow setbacks through urban planning and design guidelines and policies, so that the feeling of enclosure and appeal by the street wall can be reestablished.

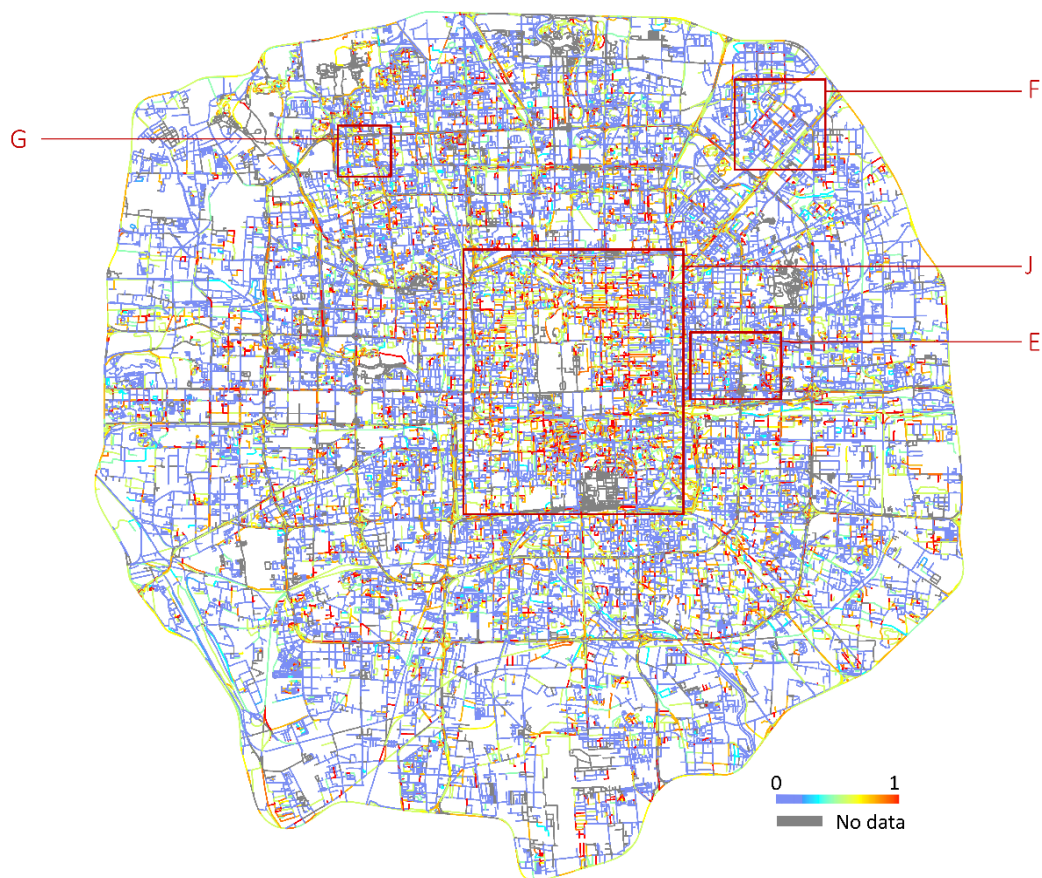

Figure 6 Visual continuity rating of street wall continuity in Beijing

## 6 Conclusion and discussion

Our aim in this paper has been to develop and test a machine learning method to automatically evaluate the urban visual environment in a large-scale. We choose two key features as the starting point of this research line, which are the visual quality of architectural façade and the visual continuity of street wall. The method can be further extended to evaluate other built environment features that shape the visual experience, such as the architectural style, the building scale, the relationship between adjacent buildings, etc.

Through applying the state-of-art deep convolutional networks, we were able to achieve a satisfying machine learning performance on the expert rated data sets. The mean squared error for the visual quality task was 0.614 on a rating scale of one to four, and the accuracy for the visual continuity task was 75%. In the next step, by conducting a field survey on the public's opinions of the built environment, we found a moderate to high correlation between the machine rating and the public's rating (Spearman's r = 0.66 for visual quality, 0.71 for visual continuity), which shows that this method produces a good approximation to the real experience in the urban environment.

Our work is also limited in several aspects. First, the size of expert labelled data set is not quite big, so we may not have achieved a maximum performance of the algorithm. To tackle this issue, we have set up a website ([www.urbanvisionstudy.com](www.urbanvisionstudy.com)) that showcases the project and advocates for crowdsourcing the labelling task so that a larger data set can be obtained. Second, although the convolutional neural network is able to capture more 'global' features (i.e. responsive to a larger region of pixel space), it may still not able to grasp all the visual cues that contribute to the judgements as mentioned in Section 3, which is still an open problem in the field of deep learning. Third, similar to the opinion of Quercia et al. (2014) that the computer algorithm is 'a tool, not a directive', we would like to say that this method provides evidence, but not decision. When it comes to the complex issue of urban planning and design, a one-size-fits-all solution does not exist and a high score in the algorithm does not always suggest the best condition. For instance, although the continuity of street wall contributes to the sense of enclosure and appeal, interruptions at certain points are also necessary to provide variety, as well as a rest for the eyes. Besides, revolutionary designs, as well as historical structures, which should be valued, may be lowly rated by the algorithm since they do not take a 'normal' look (Quercia, O'Hare et al. 2014). Therefore, it should not be oversimplified that a high scored streetscape is good enough and a low scored one is in need of change. To translate this evidence into appropriate decisions, more work is needed to understand the aesthetic cognition of the built environment through cognitive experiments, physiological psychology, etc.

This paper serves as a first-step in profiling the cityscape with computational methods. We propose that this line of research can be extended in several ways. First, as mentioned before, more urban design features can be fed into the machine learning algorithm to produce a more comprehensive profile of the urban visual environment, such as the building material, the architectural style, the building scale, the design of details, etc. Moreover, the relationship of adjacent buildings is also an important factor that shapes the streetscape, including the consistency and diversity in the use of material, color, style, scale, details, etc., as well as structural features such as the alignment of cornice, the alignment of belt course lines, etc.

Second, the long-term vision is that, with the regular update of street view images, as well as the growing volume of geo-tagged images online, we will be able to consistently monitor the transformation of the cityscape at a large scale. The urban planning issues that can only be analysed case by case on a limited data base in the past, will be easily reviewed on the city scale, e.g. 'which areas of the city are upgrading and which are decaying', 'how do new built projects complement existing buildings' geometry, scale, and/or quality of detail' (Parolek, Parolek et al. 2008), etc.

Third, cross-city and cross-regional comparison can also be an interesting direction. The cross-regional comparison is somewhat linked to the research area of *computational geo-cultural modelling* proposed by Doersch et al., which serves to provide stylistic narratives to explore the diverse visual geographies of our world (Doersch, Singh et al. 2012). Following our proposed research line, the regional differences in urban design cultures can be evaluated by comparing the aforementioned features, which may provide a deeper insight to the design cultures. Regarding to the cross-city comparison, a direct next step can be applying the algorithms developed in this paper to all the Chinese cities and produce city rankings in terms of the visual environment. In this case the primate or the most economically developed cities may not win over lower-tier cities. We expect such comparison to provide a more experience-oriented and quality-of-life-oriented perspective towards urban development, other than the measurement of hard numbers like GDP, road network density, etc.

# Reference

APA (2006). Planning and urban design standards, John Wiley & Sons.

Brownson, R. C., C. M. Hoehner, et al. (2009). "Measuring the built environment for physical activity: state of the science." American journal of preventive medicine **36**(4): S99-S123. e112.

Buchanan, P. (1988). A report from the front. Urban Design Reader. M. Carmona and S. Tiesdell, Routledge**:** 204-207.

Carmona, M. (2010). Public places, urban spaces: the dimensions of urban design, Routledge.

Devlin, K. and J. L. Nasar (1989). "The beauty and the beast: Some preliminary comparisons of 'high'versus 'popular'residential architecture and public versus architect judgments of same." Journal of Environmental Psychology **9**(4): 333-344.

Doersch, C., S. Singh, et al. (2012). "What makes Paris look like Paris?" ACM Transactions on Graphics **31**(4).

Downs, R. M. and D. Stea (1973). Image and environment: Cognitive mapping and spatial behavior, Transaction Publishers.

Ewing, R. and O. Clemente (2013). Measuring urban design: Metrics for livable places, Island Press.

Gehl, J. (2013). Cities for people, Island press.

Goel, A., M. Juneja, et al. (2012). Are buildings only instances?: exploration in architectural style categories. Proceedings of the Eighth Indian Conference on Computer Vision, Graphics and Image Processing, ACM.

Hara, K., V. Le, et al. (2013). Combining crowdsourcing and google street view to identify street-level accessibility problems. Proceedings of the SIGCHI conference on human factors in computing systems, ACM.

Harvey, C. W. (2014). "Measuring Streetscape Design for Livability Using Spatial Data and Methods."

Hwang, J. and R. J. Sampson (2014). "Divergent pathways of gentrification racial inequality and the social order of renewal in Chicago neighborhoods." American Sociological Review **79**(4): 726-751.

Kandinsky, W. (2012). Concerning the spiritual in art, Courier Corporation.

Kelly, C. M., J. S. Wilson, et al. (2013). "Using Google Street View to audit the built environment: inter-rater reliability results." Ann Behav Med **45 Suppl 1**: S108-112.

Kostof, S. (1992). The city assembled, London.

Lee, S., N. Maisonneuve, et al. (2015). Linking Past to Present: Discovering Style in Two Centuries of Architecture. IEEE International Conference on Computational Photography.

Lynch, K. (1960). The image of the city, MIT press.

Milroy, B. M. (2010). Thinking planning and urbanism, UBC Press.

Naik, N., S. D. Kominers, et al. (2015). "Do People Shape Cities, or Do Cities Shape People? The Co-evolution of Physical, Social, and Economic Change in Five Major U.S. Cities." National Bureau of Economic Research Working Paper Series **No. 21620**.

Naik, N., J. Philipoom, et al. (2014). Streetscore--predicting the perceived safety of one million streetscapes. Computer Vision and Pattern Recognition Workshops (CVPRW), 2014 IEEE Conference on, IEEE.

Ordonez, V. and T. L. Berg (2014). Learning high-level judgments of urban perception. Computer Vision–ECCV 2014, Springer**:** 494-510.

Parolek, D. G., K. Parolek, et al. (2008). Form based codes: a guide for planners, urban designers, municipalities, and developers, John Wiley & Sons.

Porzi, L., S. Rota Bulò, et al. (2015). Predicting and Understanding Urban Perception with Convolutional Neural Networks. Proceedings of the 23rd Annual ACM Conference on Multimedia Conference, ACM.

Quercia, D., N. K. O'Hare, et al. (2014). "Aesthetic capital." 945-955.

Salesses, P., K. Schechtner, et al. (2013). "The collaborative image of the city: mapping the inequality of urban perception." PLoS One **8**(7): e68400.

Shalunts, G., Y. Haxhimusa, et al. (2011). Architectural style classification of building facade windows. Advances in Visual Computing, Springer**:** 280-289.

Shalunts, G., Y. Haxhimusa, et al. (2012). Architectural style classification of domes. Advances in Visual Computing, Springer**:** 420-429.

Von Meiss, P. (2013). Elements of architecture: from form to place, Routledge.

Xu, Z., D. Tao, et al. (2014). Architectural style classification using multinomial latent logistic regression. Computer Vision–ECCV 2014, Springer**:** 600-615.